\documentclass[letterpaper, 10 pt, conference]{ieeeconf}

\makeatletter
\def\endthebibliography{%
  \def\@noitemerr{\@latex@warning{Empty `thebibliography' environment}}%
  \endlist
}
\makeatother
 
\IEEEoverridecommandlockouts
\usepackage{cite}
\usepackage{amsmath,amssymb,amsfonts}
\usepackage{graphicx}
\usepackage{textcomp}
\usepackage{xcolor}
\usepackage{todonotes}
\usepackage{amsmath}
\usepackage{leftidx}
\usepackage{siunitx}
\usepackage{multirow}
\usepackage{footnote}
\usepackage[utf8]{inputenc}
\usepackage{array}
\usepackage{mdwmath}
\usepackage{mdwtab}
\usepackage{eqparbox}
\usepackage{eufrak}
\usepackage{url}
\usepackage{cite}
\usepackage{amsmath,amssymb,amsfonts}
\usepackage{algorithm2e}
\RestyleAlgo{ruled}
\usepackage{textcomp}
\usepackage{lettrine}
\usepackage{caption}
\usepackage{subcaption}
\usepackage[export]{adjustbox}
\usepackage{amsmath}
\usepackage{leftidx}
\usepackage{siunitx}
\usepackage{multirow}
\usepackage{balance}
\usepackage[export]{adjustbox}
\usepackage{booktabs}
\usepackage{hyperref}
\usepackage{balance}

\def\BibTeX{{\rm B\kern-.05em{\sc i\kern-.025em b}\kern-.08em
    T\kern-.1667em\lower.7ex\hbox{E}\kern-.125emX}}
\begin{document}

%\title{\LARGE \bf Merging Architectural and Situational Graphs for Robot Localization}
%\title{\LARGE \bf Architectural Graphs: Unifying Situational Graphs with Architectural plans for Online Global Localization}
\title{\LARGE \bf Graph-based Global Robot Localization Informing Situational Graphs with Architectural Graphs}
%\title{\LARGE \bf Localization on Architectural Graphs}
% \title{\LARGE \bf Localization on Situational Graphs from Architectural Plans}
%\title{\LARGE \bf Localization over Hierarchical Situational Graphs Augmented using Prior Knowledge}
% Localization over Hierarchical Situational Graphs Augmented using Prior Knowledge 

\author{Muhammad Shaheer$^{1}$, Jose Andres Millan-Romera$^{1}$, Hriday Bavle$^{1}$, Jose Luis Sanchez-Lopez$^{1}$,\\ Javier Civera$^{2}$ and Holger Voos$^{1}$ 
\thanks{$^{1}$Authors are with the Automation and Robotics Research Group, Interdisciplinary Centre for Security, Reliability and Trust (SnT), University of Luxembourg (UL). Holger Voos is also associated with the Faculty of Science, Technology and Medicine, University of Luxembourg, Luxembourg.
\tt{\small{\{muhammad.shaheer, jose.millan, hriday.bavle, joseluis.sanchezlopez, holger.voos\}}@uni.lu}}% 
\thanks{$^{2}$Author is with I3A, Universidad de Zaragoza, Spain
{\tt\small jcivera@unizar.es}}%
\thanks{*This work was partially funded by the Fonds National de la Recherche of Luxembourg (FNR) under the project 17097684/RoboSAUR, by a partnership between the SnT-UL and Stugalux Construction S.A., and by the Spanish and Aragón governments (projects PID2021-127685NB-I00, TED2021-131150B-I00 and DGA\_FSE-T45\_20R).
For the purpose of Open Access, the author has applied a CC BY 4.0 public copyright license to any Author Accepted Manuscript version arising from this submission.}
}

\maketitle
\begin{abstract}

\label{abstract}
In this paper, we propose a solution for legged robot localization using architectural plans. Our specific contributions towards this goal are several. Firstly, we develop a method for converting the plan of a building into what we denote as an architectural graph (A-Graph). When the robot starts moving in an environment, we assume it has no knowledge about it, and it estimates an online situational graph representation (S-Graph) of its surroundings. We develop a novel graph-to-graph matching method, in order to relate the S-Graph estimated online from the robot sensors and the A-Graph extracted from the building plans. Note the challenge in this, as the S-Graph may show a partial view of the full A-Graph, their nodes are heterogeneous and their reference frames are different. After the matching, both graphs are aligned and merged, resulting in what we denote as an informed Situational Graph (iS-Graph), with which we achieve global robot localization and exploitation of prior knowledge from the building plans. Our experiments show that our pipeline shows a higher robustness and a significantly lower pose error than several LiDAR localization baselines.

\noindent \textbf{Paper Video:} \url{https://youtu.be/3Pv7y8aOsUY}

%In this paper, we enhance the hierarchical situational graph of environments by incorporating prior knowledge and performing real-time hierarchical localization. We introduce two new semantic entities, \textit{Wall} and \textit{Doorway}, into the factor graph. A Wall entity connects two wall surfaces, and a Doorway entity connects two rooms. Additionally, we propose a novel graph-matching-based localization approach. We construct two hierarchical representations of the environment; the first is based on prior knowledge from building architectural plans, while the second is built incrementally online as the robot navigates the environment. By matching the structure of these two graphs, we can accurately estimate the robot's location in the environment
\end{abstract}

\section{Introduction}
\label{introduction}

\begin{figure}[t]
    \centering
    \includegraphics[width=0.48\textwidth]{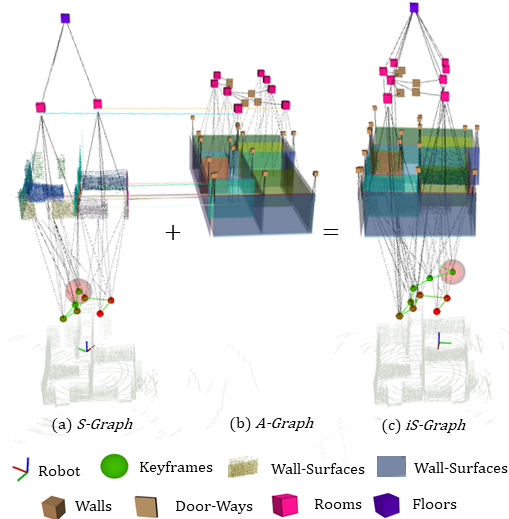}
\caption{Generation of an \textit{iS-Graph} leveraging the information from the offline generated \textit{A-Graph} using an architectural plan and the online generated \textit{S-Graph} using robot sensors. A structure-based graph matching algorithm estimates the relationship between the two graphs as the robot navigates to provide a final connected \textit{iS-Graph}.}
 \label{fig:front_image}
\end{figure}

{M}{obile} robots are increasingly being deployed in the construction sector, with significant potential benefits. For example, they may reduce significantly the costs by regular inspection of an ongoing site to monitor progress. However, robots at construction are nowadays mostly teleoperated or work semi-autonomously due among others to the perception challenges associated with the constantly changing nature of a construction site. For fully autonomous operation, it would be convenient for such robots to have a comprehensive prior knowledge of the construction site geometry. Leveraging such prior knowledge together with sensor readings during real-time operation may lead to robust and accurate global localization in construction sites. 

Digital architectural plans, such as Building Information Modelling (BIM) \cite{bim}, provide a means of capturing and  communicating information about a construction site and incorporating it as prior knowledge about the scene. Works such as \cite{connecting_semantic_bim, bim_localization, robot_localization_shaheer} have addressed the problem of extracting relevant structural knowledge from BIM and using it for real-time robot localization. However, these methods only extract geometric information from the BIM and do not leverage the topological and relational information also available in it, which limits the robustness and accuracy in complex and changing construction sites. 

To tackle this problem, we present a novel approach to localize robots leveraging not only geometry but also higher-level hierarchical information from architectural plans. We present in this paper how to model the BIM information in the form of a graph that we denote as Architectural Graph (\textit{A-Graph}), and then match and merge with the online Situational Graph (\textit{S-Graph}) \cite{s_graphs}, \cite{s_graphs+} that the robot builds as it navigates the environment. As a key aspect, translating low-level geometry into high-level features in both graphs is what allows a robust matching between such different inputs.

Our method can be divided into three main stages. In the first one, an \textit{A-Graph} is created for a given environment with nodes representing the semantic features available in a BIM model, specifically, wall-surfaces, doors, and rooms as the graph nodes, and edges containing the relevant relational information such as two wall surfaces comprising a wall, four wall surfaces connecting room and rooms connected through doors. In the second stage, running in real-time onboard the robot, a \textit{S-Graph} is estimated using 3D LiDAR measurements. The nodes of our \textit{S-Graph} correspond to semantics such as wall surfaces and rooms, and the edges correspond to constraints between these wall surfaces and the relevant room nodes. Finally, to localize the robot within this environment, a graph-matching algorithm is proposed utilizing hierarchical information from both graphs to provide the best match candidates finally resulting in informed (\textit{iS-Graphs}) that fuses the information of both. This last graph will be the one used for global localization.

As a summary, the main contributions of this paper are: 
\begin{itemize}
    \item A hierarchical optimizable \textit{A-Graph} built from BIM models that extends \textit{S-Graphs} factors~\cite{s_graphs,s_graphs+} with novel wall and door-way ones.
    \item A novel hierarchical graph matching between \textit{A-Graphs} and \textit{S-Graphs}. 
    \item Generation of \textit{iS-Graphs}, from the fusion of \textit{A-Graphs} and \textit{S-Graphs} for graph-based global robot localization.
    \item Validation of the proposed solution over simulated and real datasets collected over different construction sites achieving state-of-the-art results. 
\end{itemize}

\begin{figure*}[t]
    \centering
    \includegraphics[width=0.9\textwidth]{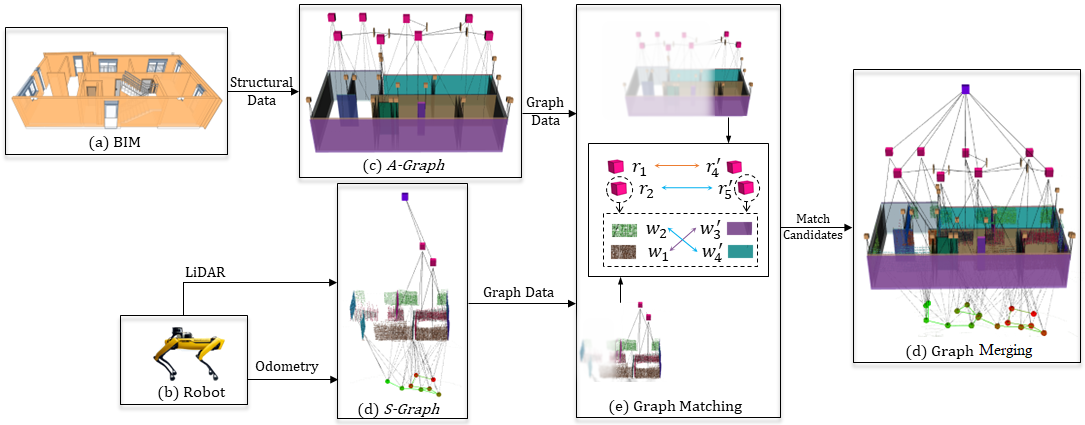}
\caption{Overview of our approach. We generate offline an Architectural Graph (\textit{A-Graph}) from a BIM model. A robot estimates online a Situational Graph (\textit{S-Graph}) from its sensors. We do graph matching between the two, align them, and merge their information. This generates the final \textit{iS-Graphs}, which is utilized by the robot to be localized with respect to the BIM.}
 \label{fig:intro_pic}
\end{figure*}

\section{Related Work}
\label{related_works}
\subsection{Global Localization}

Global localization refers to the estimation of the pose of a robot on a known map. Global localization is usually done in two steps. Initial localization is usually performed by sensor-reading-to-map matching~\cite{majdik2013mav,kim2017satellite}, which is frequently referred to as \emph{re-localization}~\cite{kendall2016modelling}, \emph{place recognition}~\cite{lowry2015visual} or \emph{the kidnapped robot prolem}~\cite{desrochers2015set}. After an initial pose has been estimated within the known map, the robot pose is tracked in such map fusing its dynamics and sensor readings.

Multiple sensor modalities, including LiDAR~\cite{miller2021any}, monocular~\cite{leung2008localization}, and stereo~\cite{zuo2020multimodal} cameras, can be employed for such tasks. Fox et al.~\cite{Markov-Loc} proposed one of the pioneer works in global localization, in which they use a Markov model to update with each sensor reading, a probability distribution over the space of potential locations in the surrounding environment. If the robot becomes confident about its location, the probability distribution converges to a single mode centered around the true position of the robot. Markov localization has, however, a high computational footprint. To overcome this issue, \cite{MCL} proposed Monte Carlo Localization (MCL), which models the robot state as a set of particles that are randomly sampled from the probability density function. MCL has proven to be effective in a wide variety of scenarios \cite{scan-MCL}, \cite{mpd-MCL}, \cite{ndt-mcl}.

Hess et al.~\cite{cartographer} proposed Cartographer, a real-time indoor mapping algorithm along with localization capability. It uses submap based scan-matching technique for estimating the initial odometry. A global loop closure is performed to optimize the estimated trajectory and further improve the accuracy of the map. The localization-only mode of Cartographer then uses this map for global localization.

Other methods rely on the extraction and matching of local point cloud descriptors such as \cite{fpfh}, \cite{sog}, \cite{bowpf}. Keypoint-based methods are constrained by the limited distinctiveness of local structures in LiDAR data. However, they have emerged as an alternative to those based on scan matching and have garnered significant interest from the research community in recent years \cite{fast_histo}, \cite{M2dp}.

Recent research has shown some promising results in learning-based localization. \cite{seg-map} presented SegMap which creates a different map representation for localization by extracting segments from 3D point clouds. Most global localization approaches assume that the prior maps were estimated using the same sensors used for localization. This limits the use of maps estimated from other sensor modalities or human maps, as we do. As we are addressing this specific last case, we refer to related works in the next subsection.

\subsection{Localization in Architectural Plans}
Boniardi et al.~\cite{PGLOC-CADfloorPlans} proposed the use of CAD plans for robust LiDAR indoor localization. They first create a globally consistent sparse map by combining Bayes filtering and graph-based mapping techniques. The robot then estimates its pose in the global frame of the floor plan by aligning this map to the CAD plan. \cite{room-layout-edge-extraction} also proposed a localization approach in digital floor plans with a monocular camera. They predict the room layout by using a convolutional neural network (CNN) named room layout extraction network. Then they use a particle filter to estimate the robot pose by matching the extracted edges from the network and the digital floor plan. \cite{GLFP} performed global localization using schematic floor plans. They proposed a factor-graph-based localization approach that uses features --such as wall intersections and corners from the digital floor plans-- as landmarks. Their approach assumes that the features mentioned above are previously labeled. The robot extracts similar features from a 3D point cloud and performs data association. The robot poses and landmarks are jointly optimized by a factor graph. \cite{Precise-Robot-Localization-in-Architectural-3D-Plans} addressed the issue of localization in erroneous architectural plans. They first segment RGB images using a segmentation network into foreground and background, then fuse them with LiDAR scans. Then they perform local referencing to make up for the deviations in the digital plan and perform precise localization. They also proposed an outlier detector to counter noise and clutter in sensor readings. They claim that their method outperforms the Iterative Closest Point (ICP) based method by almost $30 \%$. \cite{SGD-LOC} also addressed localization in inaccurate floor plans using Stochastic Gradient Descent (SGD) and 2D LiDAR. They benchmark their algorithm against Adaptive Monte Carlo Localization (AMCL) \cite{AMCL} and claim to significantly outperform it in the case of imprecise maps. Ling et al.~\cite{FP-LOC} proposed a novel idea of Approximate Nearest Neighbor Fields (ANNF) which they use for localization on floor plans. ANNF allows for fast  retrieval of the closest geometric floor plan features for any point within the localization space. \cite{Towards-BIM-based-robot-localization} proposed to use of Building Information Modelling (BIM) for robot localization. Instead of using the traditional Simultaneous Localization and Mapping (SLAM) algorithm, they convert the BIM model to a localization-oriented point cloud and localize the robot using ICP between the robot's laser scanner and the metric point cloud. \cite{vega:2022:2DLidarLocalization} proposes an open-source method to generate appropriate Pose Graph-based maps from BIM models for robust 2D-LiDAR localization in changing and dynamic environments. They first create 2D occupancy grid maps automatically from BIM (Building Information Modelling) and then convert them into Pose-Graph-based maps. Then using various map representations they perform localization using different localization algorithms. \cite{BIM-LOC} explore an effective way to localize a mobile 3D LiDAR sensor on BIM-generated maps considering both geometric and semantic properties. 

\subsection{Graph Matching}

Several approaches have tackled the problem of associating data from two different sets. ~\cite{bailey2000data} formulates a maximum common sub-graph problem, or maximum clique when new noisy data is acquired incrementally. ~\cite{antonante2021outlier} performs an outlier-robust estimation, maximum consensus, and truncated least squares. CLIPPER ~\cite{lusk2021clipper} presents a relaxation of this NP-hard problem. They weigh the geometric consistency of different associations and encounter the most consistent associations (inliers) by finding the densest sub-graph, hence as a graph theoretic framework. The authors provide experiments with point and point-normal noisy sets with outliers. ~\cite{arandjelovic2016netvlad} use a convolutional neural network for large places recognition visually. 

The literature has approached the problem of graph matching for robot localization with a loop-closure goal. ~\cite{global_loc_gnn}  takes as input two graphs describing scenes. Random walks are accomplished in both of them generating class-aware descriptors of each walk. The node association is based on the number of identical descriptors in common. Distance comparison between nodes is used when ambiguity remains for different match candidates.~\cite{semantic_loop_closure} use a sparse Kuhn–Munkres algorithm to find the association between nodes of graphs extracted from images. The similarity is computed in terms of shape and image similitudes along the euclidian distance. ~\cite{zheng2020buildingfusion} measure the similarities by comparing the instance-level embeddings provided by 3D CNNs followed by a graph-wise geometrical verification.

The concept of 3D scene graphs for representing the structure of an environment in a hierarchical manner was introduced by \cite{scene_graph} and \cite{3d_scene_graph}.
~\cite{hydra} presented a hierarchical loop-closer detection in graphs. First, they compare descriptors in a top-down fashion to find putative loop closures at places, objects, and vision. Those descriptors contain statistics of the robot’s surroundings at object and place levels combining geometrical and semantic information. Both are aggregated as histograms for its comparison. Second, a bottom-up verification of the geometric consistency yields a match between robot nodes.
\section{System Overview}
\label{system_overview}
Fig.~\ref{fig:intro_pic} shows an overview of the proposed approach. Firstly the BIM information is extracted for a given environment to create the two-layered \textbf{Architectural Graphs} (Section.~\ref{sec:a_graphs}) in an offline manner, with all the elements extracted in the BIM frame of reference $B$. As the robot navigates within the given environment, an online \textbf{Situational Graph} (Section.~\ref{subsec:situational_graphs}) is estimated by the robot in the map frame of reference $M$. In parallel, we run our the \textbf{Graph Matching} method (Section.~\ref{sec:graph_matching}) to provide match candidates between the existing \textit{A-Graphs} and the current \textit{S-Graph}. Finally, after retrieving the best match candidate, our \textbf{Graph Merging} (Section.~\ref{sec:graph_merging}) provides the merged \textit{iS-Graph} utilized for the global localization with respect to the BIM frame of reference $B$.

% can be divided into four main modules: 
% \textbf{Architectural Graphs}, \textbf{Situational Graphs}, \textbf{Graph Matching}, and \textbf{Graph Merging}. The architectural graph is extracted in the frame of reference defined in BIM called $B$. The \textit{S-Graph} is built online in the map frame of reference $M$. The final merged graph is with respect to the BIM frame of reference $B$.

\subsection{Situational Graphs (S-Graphs)} \label{subsec:situational_graphs}

\textit{S-Graphs} are four-layered optimizable hierarchical graphs built online using 3D LiDAR measurements. The full details of the \textit{S-Graphs} we use in this work can be found in \cite{s_graphs},\cite{s_graphs+}. In brief, their four layers can be summarized as:

\textbf{Keyframes Layer.} It consists of the robot poses factored as $\leftidx{^M}{\mathbf{x}}_{R_i} \in SE(3)$ nodes in the map frame $M$ with pairwise odometry measurements constraining them. 

\textbf{Walls Layer.} It consists of the planar wall-surfaces $\leftidx{^M}{\boldsymbol{\pi}}_{i} \in \mathbb{R}^3$ extracted from the 3D LiDAR measurements and factored using minimal plane parameterization. The planes observed by their respective keyframes are factored using pose-plane constraints. 

\textbf{Rooms Layer:} It consists of two-wall rooms $\leftidx{^M}{\boldsymbol{\gamma}}_{i} \in \mathbb{R}^2$ or four-wall rooms $\leftidx{^M}{\boldsymbol{\rho}}_{i} \in \mathbb{R}^2$, each constraining either two or four detected wall-surfaces respectively. 

\textbf{Floors Layer:} It consists of a floor node $\leftidx{^M}{\boldsymbol{\xi}}_{i} \in \mathbb{R}^2$ positioned at the center of the current floor level and constraining all the rooms present at that floor level.

%TODO:HB add the state vector description here

\section{Architectural Graphs (A-Graphs)} \label{sec:a_graphs}
% Fig.~\ref{fig:system_architecture}.  

We propose to extract relevant information from BIM models into two-layered optimizable graphs denoted as \textit{A-Graphs}. In the lowest-level layer we will model the geometry of the walls, and in the highest level the rooms. Room-to-walls constraints connect the two layers and neighbouring rooms are constrained by door-ways. The specific formulation is detailed below.

\textbf{A-Walls Layer.} \label{sec:pose_graph_gen} This layer extracts all the information about the walls and wall surfaces from the BIM and connects them with appropriate wall-to-wall-surface edges. 

\textbf{\textit{Wall-Surfaces:}} Wall-surfaces are planar entities $\leftidx{^{B}}{\boldsymbol{\pi}}$ extracted in the BIM frame of reference $B$. All the wall-surfaces are converted to their Closest Point (CP) representation, as in \cite{s_graphs+}. Wall-surface normals with their component $\leftidx{^B}{{n}_x}$ greater than $\leftidx{^B}{{n}_y}$ are classified as $x$-wall-surfaces, and wall surfaces whose normal component $\leftidx{^B}{{n}_y}$ is greater than the normal component   $\leftidx{^B}{{n}_x}$ are defined as $y$-wall-surfaces. 
These are initialized in the graph as $\leftidx{^B}{\boldsymbol{\pi}} = [\leftidx{^B}\phi, \leftidx{^B}\theta, \leftidx{^B}d]$, where $\leftidx{^B}\phi$ and $\leftidx{^B}\theta$ stand for the azimuth and elevation of the plane in frame $B$ and $\leftidx{^B}d$ is the perpendicular distance in $B$. 

\textbf{\textit{Walls:}} 
We introduce a novel semantic entity with respect to \cite{s_graphs+} in the form of a \textit{Wall} $\boldsymbol{\omega} \in \mathbb{R}^3$, consisting of two planar wall-surfaces. Two opposed planar wall-surface entities either in $x$-direction or $y$-direction with similar perpendicular distance $\leftidx{^B}d$ can be classified as a part of a single wall entity. The wall center $\leftidx{^B}{\boldsymbol{\omega}_{x_i}}$ for two opposed $x$-direction wall-surfaces is computed as:

\begin{gather}
\resizebox{1.\hsize}{!}{$\leftidx{^B}{\mathbf{w}_{x_i}} =  \frac{1}{2} 
 \big[ \lvert {\leftidx{^B}{d_{x_{1}}} \rvert} \cdot \leftidx{^B}{\mathbf{n}_{x_{1}}} - {\lvert \leftidx{^B}{d_{x_{1}}} \rvert} \cdot \leftidx{^B}{\mathbf{n}_{x_{2}}} \big]  + \lvert \leftidx{^B}{{d_{x_{2}}} \rvert} \cdot \leftidx{^B}{\mathbf{n}_{x_{2}}} \nonumber$} \\
 \leftidx{^B}{\boldsymbol{\omega}_{x_i}} = \leftidx{^B}{\mathbf{w}_{x_i}} +  \big[  \leftidx{^B}{\mathbf{\mathfrak{s}}_i} - [\ \leftidx{^B}{\mathbf{\mathfrak{s}}_i} \cdot \leftidx{^B}{\hat{\mathbf{w}}_{x_i}} ] \ \cdot \hat{\leftidx{^B}{\mathbf{w}}_{x_i}} \big]
\end{gather} \label{eq:wall_center}

\noindent where $\leftidx{^B}{\mathbf{\mathfrak{s}}_i} \in \mathbb{R}^3$ is the starting point for a given BIM wall and $\mathbf{n}$ and $d$ are the plane normals and distance. For Eq.~\ref{eq:wall_center} to hold true, all plane normals are converted to point away from the BIM frame of reference as in \cite{s_graphs+}. The wall center along with it wall-surfaces is factored in the graph as:

\begin{multline} \label{eq:infinite_room_node}
    c_{\boldsymbol{\omega}}(\leftidx{^B}{\boldsymbol{\omega}_i},\big[\leftidx{^B}{\boldsymbol{\pi}_{x_{1}}}, \leftidx{^B}{\boldsymbol{\pi}_{x_{1}}}, \leftidx{^B}{\mathfrak{s}}_i]) \\ = \sum_{i=1}^{K} \| \leftidx{^B}{\hat{\boldsymbol{\omega}}_i} - f(\leftidx{^B}{\tilde{\boldsymbol{\pi}}_{x_{1}}}, \leftidx{^B}{\tilde{\boldsymbol{\pi}}_{x_{1}}}, \leftidx{^B}{\mathfrak{s}_i}) \| ^2_{\mathbf{\Lambda}_{\boldsymbol{\tilde{\boldsymbol{\omega}}}_{i,t}}}
\end{multline}

Where $f(\leftidx{^B}{\tilde{\boldsymbol{\pi}}_{x_{1}}}, \leftidx{^B}{\tilde{\boldsymbol{\pi}}_{x_{1}}}, \leftidx{^B}{\mathfrak{s}_i})$ is the function mapping the wall center using the wall-surfaces and its starting point following Eq.~\ref{eq:wall_center}. Wall factors add an additional layer of structural consistency to the graph. A Wall center for opposed planes in $y$-direction is computed following Eq.~\ref{eq:wall_center}.

\textbf{A-Rooms Layer.} The second layer of the graph extracts all the information about the rooms along with the door-ways interconnecting the rooms.   

\textit{\textbf{Rooms:}} In this work we use the similar concept of a four-wall room $\leftidx{^B}{\boldsymbol{\rho}} \in \mathbb{R}^2$ as presented in \cite{s_graphs+}, where each room comprises the four-wall surfaces extracted in the first layer of the graph. The information regarding the room with its wall surfaces can be easily extracted from the BIM. The creation of the room and room-to-wall-surfaces edges can be referred to in \cite{s_graphs+}.

\textit{\textbf{Door-Ways:}} In this paper we incorporate an additional entity in the graph called door-ways interconnecting the room nodes, easily available from BIM. The position of a door-way node $\leftidx{^B}{\boldsymbol{\mathcal{{D}}}} \in \mathbb{R}^3$ is directly extracted from BIM in the frame of reference $B$. Using the semantic information from BIM of the rooms connected by a given door-way, the door-way-to-rooms factor can be formulated as:
\begin{multline}
c_{\boldsymbol{\mathcal{D}}} (\leftidx{^B}{\boldsymbol{\rho}_1},\leftidx{^B}{\boldsymbol{\rho}_2} , \leftidx{^B}{\boldsymbol{\mathcal{D}}_{i}}) = \\ \| \ f(\leftidx{^{B}}{\hat{\boldsymbol{\rho}}_{1}},
\leftidx{^{\rho_1}}{\hat{\boldsymbol{{\mathcal{D}}}}_i})
- \ f(\leftidx{^B}{\hat{\boldsymbol{\rho}}_2}, \leftidx{^{\rho_2}}{\hat{\boldsymbol{\mathcal{D}}}_i}) \|
\end{multline}

Where $\leftidx{^B}{\boldsymbol{\rho}_1}$ and $\leftidx{^B}{\boldsymbol{\rho}_2}$ are the four-wall rooms connected to the door-way $\leftidx{^B}{\boldsymbol{\mathcal{D}}_{i}}$. $\leftidx{^{\rho_1}}{\boldsymbol{\mathcal{D}}_i}$ and $\leftidx{^{\rho_2}}{\boldsymbol{\mathcal{D}}_i}$ are the positions the door-way nodes estimated with respect to rooms $\leftidx{^B}{\boldsymbol{\rho}_1}$ and $\leftidx{^B}{\boldsymbol{\rho}_2}$.

\section{Graph Matching}
\label{sec:graph_matching}

\begin{figure*}[t]
    \centering
    \includegraphics[width=0.9\textwidth]{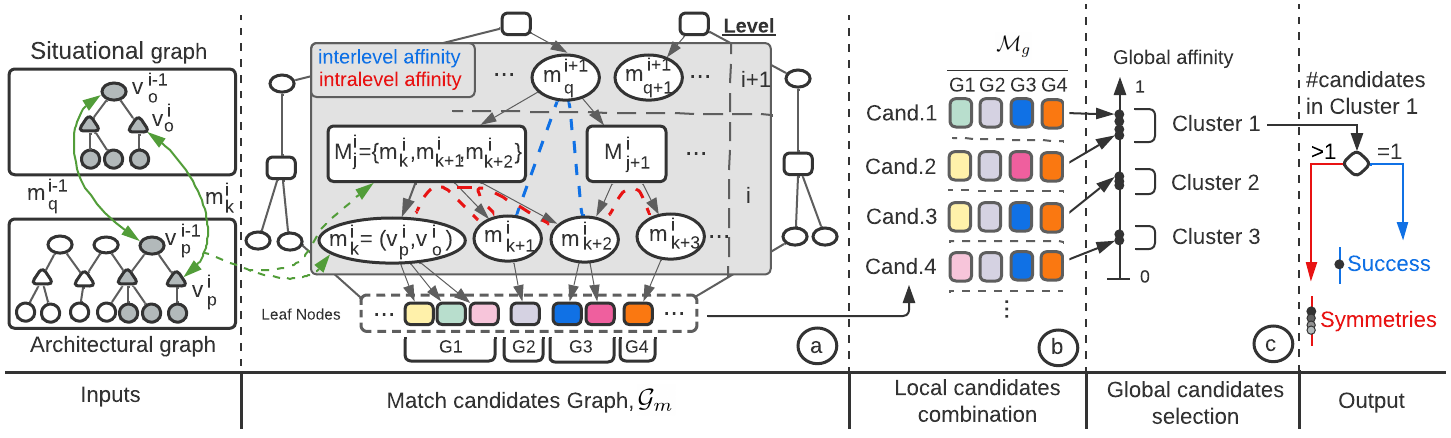}
    \caption{Graph matching schema. a) Downwards and at each level, different combinations of matches are proposed and selected by their geometrical affinity either at the same level or with the associated upper-level pair. b) The match graph is traversed upwards while combined with same-level nodes to define all-level match candidates. c) The lowest-level pairs of every candidate are scored in global affinity. That score is clustered to find symmetries in the best cluster.}
    \label{fig:graph_match_schema}
    %\vspace{-2mm}
\end{figure*}

Our second contribution to this paper is a novel approach to graph matching. Specifically, our goal is to match an architectural \textit{A-graph}, $\mathcal{G}_a$, and a \textit{S-Graph}, $\mathcal{G}_s$. In our case, we compare and match the room ($\rho$) entities and for each room entity its corresponding wall-surfaces ($\pi$), extracted from both graphs. As a result, the correspondences form a bipartite graph connecting the nodes of some parts of $\mathcal{G}_a$ with all, or almost all nodes, in $\mathcal{G}_s$, at the rooms and walls layers. 
Provided that $\mathcal{G}_s$ is built incrementally as the robot navigates the environment, the graph matching is run after every map update until a successful match is obtained. 
%At that moment we can hence determine which parts of the building have been traversed so far by the robot. 
A schema of the entire graph-matching process is described in Fig.~\ref{fig:graph_match_schema}.

\textbf{Notation.} Let $\boldsymbol{V}$ be any set of nodes, with $\boldsymbol{V}_a$ and $\boldsymbol{V}_s$ the sets of nodes in the \textit{A-Graph} and the \textit{S-Graph} respectively. Let $\boldsymbol{m} = {(v_a, v_s) : v_a \in V_a, v_s \in V_s}$, and $\boldsymbol{M}$ be any set of $m$ such as $\boldsymbol{M} =  \{\boldsymbol{m}_1, \boldsymbol{m}_2, ..., \boldsymbol{m}_n\}$. Let $\boldsymbol{\mathcal{M}}$ be any set of  $\boldsymbol{M}$ such as $\boldsymbol{\mathcal{M}} = \{\boldsymbol{M}_1, \boldsymbol{M}_2, ..., \boldsymbol{M}_n\}$. Local candidates are $\boldsymbol{M}$ including $\boldsymbol{m}$ referring to a small part of the input graphs. Global candidates are $\boldsymbol{M}$ including $\boldsymbol{m}$ referring to the entire input graphs. 

%\subsection{Match candidates graph definition}

\textbf{Match Candidates Graph.} The first process is to build a graph $\mathcal{G}_m$ with the match candidates that, incrementally and in a top-down manner, integrates and manages all suitable match combinations at each level as described in Fig.~\ref{fig:graph_match_schema}a.  
Level $i$ comprises two sub-levels. At the lowest sub-level, $\boldsymbol{m}^i_{k}$, $k=\{1,\hdots n\}$ contains a single edge linking a node in $\mathcal{G}_a$ with another node in $\mathcal{G}_s$. The second sub-level, $\boldsymbol{M}^i_{j}$, $j=\{1,\hdots n\}$ contains a suitable combination of $\boldsymbol{m}^i_{k}$ elements. 
%Notice that different combinations $\boldsymbol{M}^i_{j}$ could include the same element $\boldsymbol{m}^i_k$. 
In this work, $\boldsymbol{m}^i_k$ are pairs of room nodes and their leaf node contains the pairs of wall-surfaces for a given room node pair.  

The suitability of each $\boldsymbol{M}^i_j$ candidate is checked by three different factors required to be satisfied. First, we check isomorphism, i.e., the structure of both matched sub-graphs must be identical. Second, we perform a categorical check, i.e., the type of two matched nodes must be the same type. And third, affinity is checked, i.e. both matched sub-graphs must maintain geometrical consistency with respect to all other nodes either at the same level (interlevel) or with other levels (interlevel). The geometrical consistency is computed as defined in ~\cite{lusk2021clipper} for rooms (points) and wall surfaces (point-normals).

\textbf{Local Candidates Combination.} 
%In order to maintain global affinity, different branches must be assessed to be geometrically consistent.
In the bottom layer of $\mathcal{G}_m$, every leaf node represents a partial match for a single branch of the graph. For that purpose, as described in Fig.~\ref{fig:graph_match_schema}b, all branch combinations of $\mathcal{G}_m$ are composed of different candidates. In a bottom-up manner for every level, children of $\boldsymbol{M}$ type nodes are inclusively aggregated and children $\boldsymbol{m}$ type nodes are exclusively aggregated. As a result, $\boldsymbol{\mathcal{M}}_g$ is obtained, represented in the figure only by leaf nodes. The global affinity of the leaf nodes in each element of $\boldsymbol{\mathcal{M}}_g$ is then ranked to determine which combinations are globally affine.

%\begin{algorithm}[h]
%\caption{Local candidates combination}\label{alg:data_association_algo2}
%\textbf{Input}: Match graph $G_{match}$, swept levels $lvls$  \\
%\textbf{Output}: List of full-graph candidates $\mathcal{M}_{f.c.}$ \\
%\hrulefill \\
%$\mathcal{M}_{f.c.} \gets CLMC(G_{match}, [], 0)$\\
%\hrulefill \\
%\SetKwFunction{FMain}{$CLMC(G^{local}_{match}, M_{to\_leaf}, i)$}
%\SetKwProg{Fn}{Function}{:}{\KwRet}
%  \Fn{\FMain}{
%    $\mathcal{M} = G^{local}_{match}.find\_nodes(lvls[i])$\\
%    \For{$M^{i}_{j} \in \boldsymbol{\mathcal{M}}$}{
%        $M_{to\_leaf}.append(M^{i}_{j})$\\
%        $\mathcal{M}_{cand} \gets []$\\
%        \eIf{$i + 1 < len(lvls)$}{
%            $M_{neigh} = G^{local}_{match}.neigh(M^{i}_{j}, lvls[i])$\\
%            $\mathcal{M}_M \gets []$\\
%            \For{$m \in \boldsymbol{M_{neigh}}$}{
%                $G_{neigh}^{single} \gets G_{match}.neigh(m^{i}_{k}, lvls[i+1])$\\
%                $\mathcal{M}_m \gets CLMC(G_{neigh}^{single}, M_{to\_leaf}, lvls[i+1])$\\
%                $\mathcal{M}_M.combine_{excl}(\mathcal{M}_m)$\\
%            }
%        }{
%            $\mathcal{M}_M \gets M_{to\_leaf}$\\
%        }
%        $\mathcal{M}_{cand}.combine_{incl}(\mathcal{M}_M)$
%    }
%    %\Return $\mathcal{M}_{cand}$
%    }    
%    \KwRet $y$\
%\end{algorithm}

%\subsection{Global candidates selection}

\textbf{Global Candidates Selection.} Due to existing symmetries, only selecting the element of $\boldsymbol{\mathcal{M}}_g$ with a higher affinity score is not enough, as there may be a set of optimal matchings and not a unique one. As an illustrative example, let $\mathcal{G}_a$ have two square rooms with similar but not identical layouts. Let $\mathcal{G}_s$ have a single room corresponding to one of those in $\mathcal{G}_a$. This case would generate a set of affinities such as two clusters 1 and 2 in Fig.~\ref{fig:graph_match_schema}c, created due to similarities between the rooms as well as similarities between the wall-surfaces of those rooms.  
%Due to the similarity between rooms, we would obtain the two clusters and due to wall-surface symmetry, we would obtain four elements in each cluster. 
%The clustering algorithm used is Density-Based Spatial Clustering of Applications with Noise (DBACAN).
For this reason, after clustering, the best cluster (cluster 1) has to contain a single set of matches to determine a unique optimal match. In the case of several optimal matches within the cluster, the algorithm awaits further information to be incorporated in $\mathcal{G}_s$ (additional rooms and wall-surfaces) before providing a single unique match.

%\begin{algorithm}[h]
%\caption{Best full-graph candidates selection}\label{alg:data_association_algo3}
%\textbf{Input}: List of full-graph candidates $\mathcal{M}_{f.c.}$ \\
%\textbf{Output}: List of affine match candidates $\mathcal{M}_{a.c.}$ \\
%$\mathcal{M}_{last} \gets \mathcal{M}_{cand}.find\_nodes(lvls.last)$\\
%$\mathcal{M}_{best} \gets []$\\
%\For{$M \in \boldsymbol{\mathcal{M}_{last}}$}{
%   $score \gets aff_{intralvl}(M)$\\
%    \If{$score > thr^{final}_{intralevel}$}{
%       $\mathcal{M}_{best}.append(M, score)$\\
%   }
%
%\mathcal{M}_{a.c.} = \mathcal{M}_{best}[cluster(\mathcal{M}_{best}.scores).best]]$\\

%\end{algorithm}

\begin{figure*}[t]
\centering
\begin{subfigure}{0.23\textwidth}
\centering
\includegraphics[width=1\textwidth]{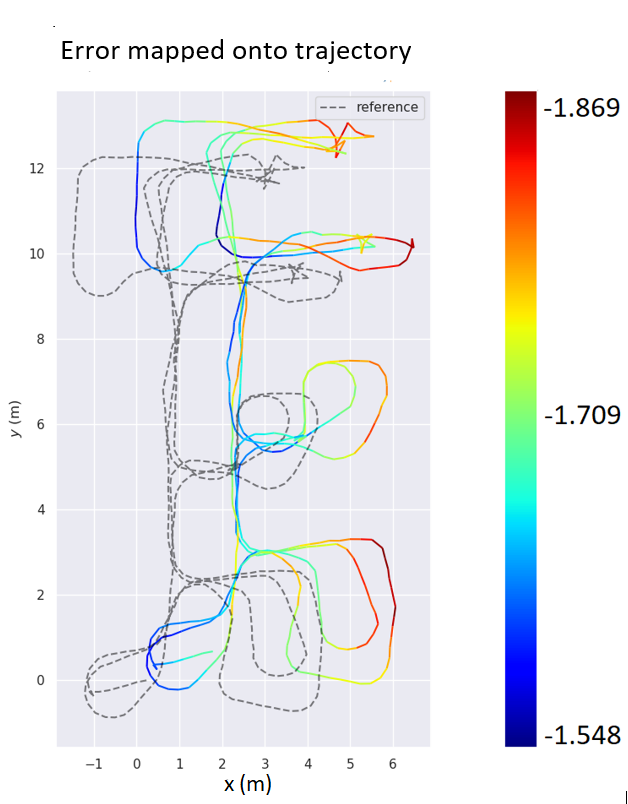}
\caption{AMCL}
\label{fig:ape_amcl}
\end{subfigure}
\begin{subfigure}{0.23\textwidth}
\centering
\includegraphics[width=0.86\textwidth]{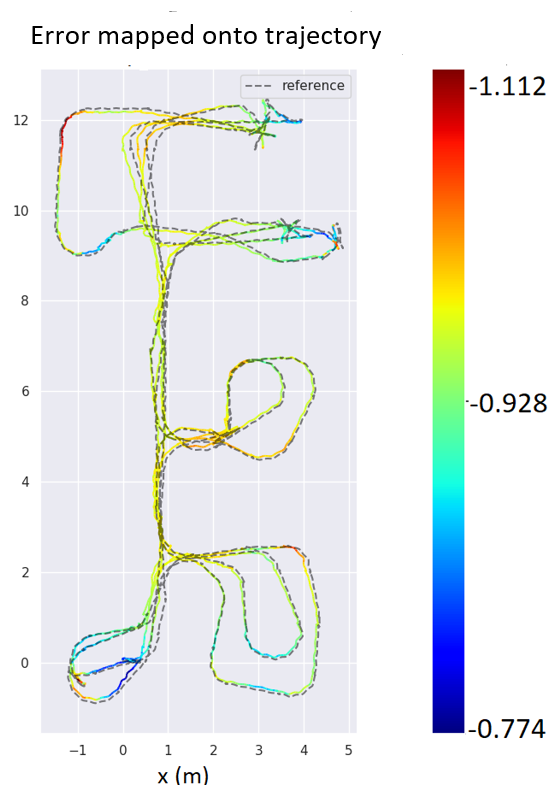}
\caption{UKFL}
\label{fig:ape_ukfl}
\end{subfigure}
\begin{subfigure}{0.23\textwidth}
\centering
\includegraphics[width=0.9\textwidth]{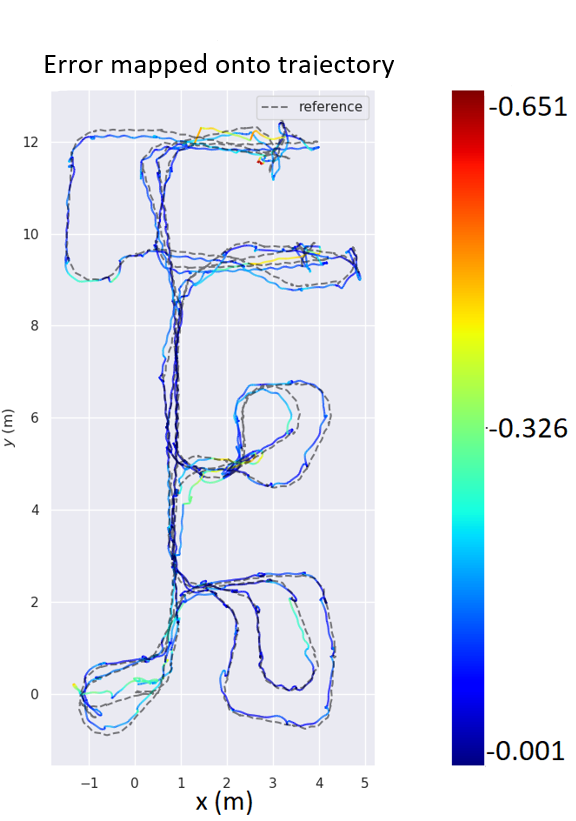}
\caption{Cartographer}
\end{subfigure}
\begin{subfigure}{0.23\textwidth}
\centering
\includegraphics[width=0.9\textwidth]{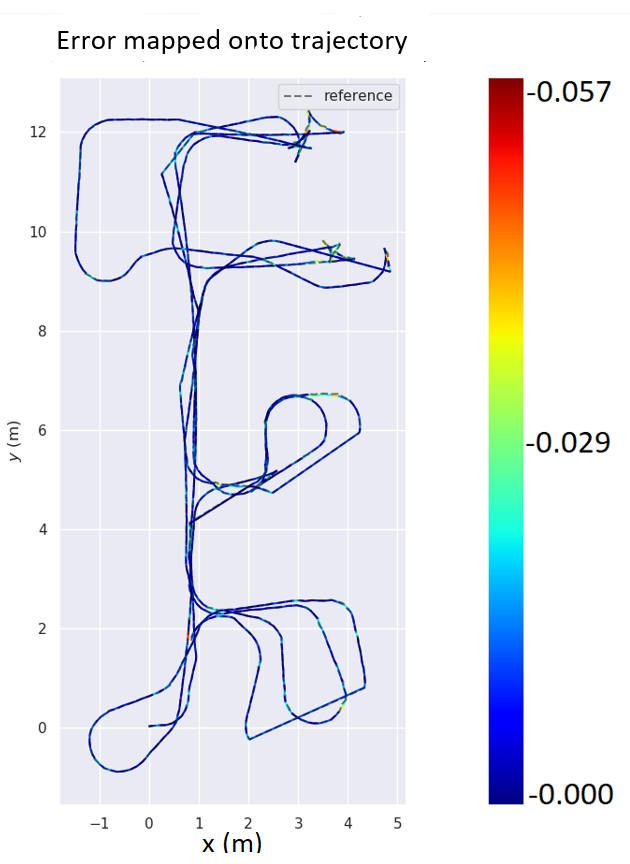}
\caption{\textit{iS-Graphs}}
\label{fig:ape_carto}
\end{subfigure}
\label{fig:3d_map_ukfl}
\caption{Top view of estimated trajectories and APE [m] for all baselines and our \textit{iS-Graph} in the D2 sequence of our simulated data. Our \textit{iS-Graph} presents the lowest errors (see the color coding), followed by Cartographer.}
\label{fig:ape_simulation}
\end{figure*}

% \begin{figure*}[t]
% \centering
% \begin{subfigure}{0.23\textwidth}
% \centering
% \includegraphics[width=1.04\textwidth]{Figures/amcl_xy_plot.png}
% \caption{AMCL}
% \label{fig:pf}
% \end{subfigure}
% \rulesep
% \begin{subfigure}{0.23\textwidth}
% \centering
% \includegraphics[width=0.9\textwidth]{Figures/hdl_xy_plot.png}
% \caption{UKFL}
% \end{subfigure}
% \rulesep
% \begin{subfigure}{0.23\textwidth}
% \centering
% \includegraphics[width=0.95\textwidth]{Figures/cartographer_xy_plot.png}
% \caption{Cartographer}
% \end{subfigure}
% \rulesep
% \begin{subfigure}{0.23\textwidth}
% \centering
% \includegraphics[width=0.93\textwidth]{Figures/isgraph_xy.png}
% \caption{\textit{iS-Graph}}
% \end{subfigure}
% \label{fig:}
% \caption{Top view of estimated trajectories and APE [m] for all baselines and our \textit{iS-Graph} in the D2 sequence of our simulated data. Our \textit{iS-Graph} presents the lowest errors (see the color coding), followed by Cartographer.}
% \label{fig:ape_simulation}
% \end{figure*}

\section{Graph Merging}
\label{sec:graph_merging}

Our global state $\mathbf{s}$ at time $T$, before graph merging, contains all the nodes of the \textit{A-Graph}, generated offline, as well as the current nodes estimated online by the \textit{S-Graph}

\begin{eqnarray}
\mathbf{s} &=& [\leftidx{^M}{\mathbf{x}}_{R_1}, \ \hdots, \ \leftidx{^M}{\mathbf{x}}_{R_T}, \nonumber \\
 & & \leftidx{^M}{\boldsymbol{\pi}}_{1}, \ \hdots, \ \leftidx{^M}{\boldsymbol{\pi}}_{P}, \ \leftidx{^B}{\boldsymbol{\pi}}_{1}, \ \hdots, \ \leftidx{^B}{\boldsymbol{\pi}}_{Q}, \nonumber \\
& & \leftidx{^M}{\boldsymbol{\rho}}_{1}, \ \hdots, \ \leftidx{^M}{\boldsymbol{\rho}}_{S}, \ \leftidx{^B}{\boldsymbol{\rho}}_{1}, \ \hdots, \ \leftidx{^B}{\boldsymbol{\rho}}_{R} \\
& & \leftidx{^M}{\boldsymbol{\gamma}}_{1}, \ \hdots, \  \leftidx{^M}{\boldsymbol{\gamma}}_{G},   
\leftidx{^M}{\boldsymbol{\xi}}_{1}, \ \hdots, \ \leftidx{^M}{\boldsymbol{\xi}}_{E}, \nonumber \\
& & \leftidx{^B}{\boldsymbol{\omega}_{1}}, \ \hdots, \  \leftidx{^B}{\boldsymbol{\omega}_{W}}, \  \leftidx{^B}{\boldsymbol{\mathcal{D}}}_{1}, \ \hdots, \  \leftidx{^B}{\boldsymbol{\mathcal{D}}}_{D}, \nonumber
\\
& & \leftidx{^B}{\mathbf{x}}_{M}]^\top, \nonumber
\end{eqnarray}

\noindent where $\leftidx{^B}{\mathbf{x}}_{M}$ is the estimated transformation between the map frame $M$ of the \textit{S-Graph} and the BIM frame \textit{B} of the \textit{A-Graph}, which is set to identity before graph merging. 
The graph matching method from Section.~\ref{sec:graph_matching} provides match candidates between the room nodes and the wall-surface nodes of the \textit{S-Graph} and the \textit{A-Graph}. To efficiently merge the two graphs to generate the \textit{iS-Graph}, we introduce room-to-room constraints as well as a wall-surface-to-wall-surface constraints between the matched candidates. The room-to-room constraint is defined as

\begin{equation}
  c_{\boldsymbol{\rho}} (\leftidx{^B}{\boldsymbol{\rho}_{1}},\leftidx{^M}{\boldsymbol{\rho}_{2}}) = \| \ \leftidx{^B}{\hat{\boldsymbol{\rho}}_{{1}}} - \leftidx{^M}{\hat{\boldsymbol{\rho}}_{{2}}} \|^2_{\mathbf{\Lambda}_{\boldsymbol{\tilde{\boldsymbol{\rho}}}_{1,2}}}~,
\end{equation} 

\noindent where $\leftidx{^B}{\boldsymbol{\rho}_{1}}$ is the room node in the \textit{A-Graph} and $\leftidx{^M}{\boldsymbol{\rho}_{2}}$ is the corresponding room node in \textit{S-Graph}. Similarly, for all correspondences between wall-surface candidates, the constraint is formulated as

\begin{equation}
  c_{\boldsymbol{\pi}} (\leftidx{^B}{\boldsymbol{\pi}_{1}},\leftidx{^M}{\boldsymbol{\pi}_{2}}) = \| \ \leftidx{^B}{\boldsymbol{\pi}_{1}} - \leftidx{^M}{\boldsymbol{\pi}_{2}} \|^2_{\mathbf{\Lambda}_{\boldsymbol{\tilde{\boldsymbol{\pi}}}_{1,2}}}~,
\end{equation}

\noindent where $\leftidx{^B}{\boldsymbol{\pi}_{1}}$ and $\leftidx{^M}{\boldsymbol{\pi}_{2}}$ are the wall-surfaces in the \textit{A-Graph} and the \textit{S-Graph}. 

With the constraints between the two graphs, $\leftidx{^B}{\mathbf{x}}_{M}$ can be estimated and all the robot poses, the wall-surfaces, rooms, and floors of the \emph{S-Graph} can be referred accurately with respect to the BIM frame of reference $B$ of the \emph{A-Graph}, resulting in the final improved situational graph \textit{iS-Graph}. In this manner the robot is localized with respect to the global reference of the architectural plan. Our approach thus can perform global localization exploiting the hierarchical high-level information in the environment without the need for appearance-based loop closure constraints at keyframe level, which are more variable.

\section{Experimental Evaluation}
\label{experimental_evaluation}
%\todo{add graph matching setup}
\subsection{Experimental Setup}
We conducted experiments to validate our proposed approach in multiple simulated and real-world construction environments. 
As it is evident, in cases where the environment has high symmetry, the localization algorithm will struggle to find a solution. Therefore, in our environments, there is always at least one unique combination of rooms.
In all experiments, we generated \textit{A-Graphs} from various building models that were created in Autodesk Revit\footnote{\url{https://www.autodesk.com/products/revit/architecture}}. The Boston Dynamics \textit{Spot} robot, equipped with a Velodyne VLP-16 3D LiDAR, was teleoperated for data collection over the environments. 
%The robot was  in both simulated environments and real construction sites. 
We compared our approach to two 2D LiDAR-based localization algorithms (AMCL \cite{AMCL} and Cartographer \cite{cartographer}) and one 3D LiDAR-based localization (UKFL \cite{hdl_graph_slam}) algorithm. In simulated datasets we measured the Absolute Pose Error (APE) \cite{evo_traj_calc} with respect to the ground truth, while in real-world due to the absence of ground truth pose information, we compare the point cloud RMSE of the generated map with the available ground truth map from BIM. 
The proposed methodology was implemented in C++, and the experiments were validated on an Intel i9 16-core workstation.

\begin{table}[b]
\centering
\caption{Absolute Pose Error (APE) [m] for  several LiDAR-based localization baselines and our \textit{iS-Graphs} Datasets have been recorded in simulated environments. `$-$' stands for localization failure.}
\begin{tabular}{c|c c c c c c}
\toprule
\textbf{Method} &  & \multicolumn{4}{c}{\textbf{APE [m] $\downarrow$}}   \\
\toprule
 & & \multicolumn{4}{c}{\textbf{Datasets}} \\ 
 \midrule
  & D1 & D2 & D3 & D4 & D5 & D6   \\ 
\midrule
AMCL \cite{AMCL} & 2.04 & 1.71 & 2.03 & 2.01 & - & -   \\ 
UKFL \cite{hdl_graph_slam} &  0.97 & 0.78 & - & 0.74 & 0.70 & 0.88 \\
Cartographer \cite{cartographer} &   0.10 & 0.16 & 0.12 & - & 0.13 & -\\
\textit{iS-Graphs (ours)} &  \textbf{0.08} & \textbf{0.01} & \textbf{0.02} & \textbf{0.04} &   \textbf{0.09} &   \textbf{0.12} \\
\bottomrule
\end{tabular}
\label{tab:ape_simulation}
\end{table}

\begin{table}[b] %pcl_compute_cloud_error source.pcd target.pcd output_intensity.pcd <options>
\setlength{\tabcolsep}{4pt}
\caption{Point cloud RMSE [m] for our real-world dataset. Best results are  boldfaced. `$-$' stands for localization failure.}
\small
\centering
\begin{tabular}{l | c c c c c}
\toprule 
\textbf{Method} & \multicolumn{3}{l}{\textbf{Alignment Error $\downarrow$}} & & \\
\toprule
  &  & \textbf{ Datasets}  &    \\ 
\midrule
  & {D1} &  {D2}  & {D3}   \\ 
\midrule
AMCL \cite{AMCL}  &  - & 0.90 &  0.98  \\ 
UKFL \cite{hdl_graph_slam}  & - & 0.86 & 0.69 \\ 
Cartographer \cite{cartographer} & - & 0.58 & 0.64  \\
% \midrule
\textit{iS-Graphs (ours)} & \textbf{0.17} & \textbf{0.20} & \textbf{0.21}   \\ 
\bottomrule 
\end{tabular}
%\vspace{-3mm}
\label{tab:pcerror_real_data}
\end{table}
\label{ex_grapgh_gen}

\begin{figure*}[h]
\centering
\begin{subfigure}{0.18\textwidth}
\centering
\includegraphics[width=0.75\textwidth]{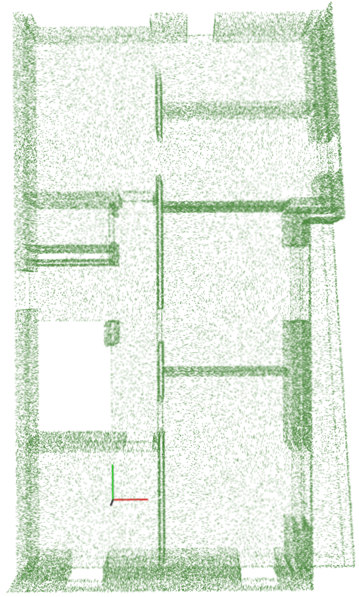}
\caption{BIM}
\label{fig:3d_map_gt}
\end{subfigure}
\begin{subfigure}{0.18\textwidth}
\centering
\includegraphics[width=0.80\textwidth]{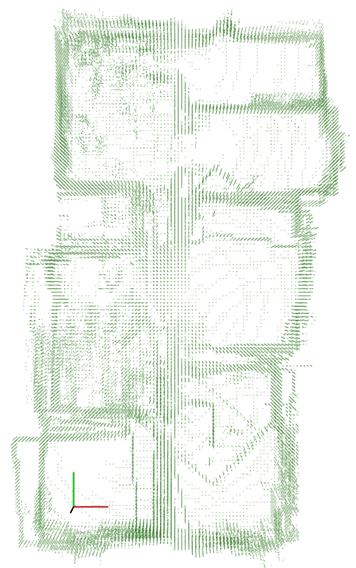}
\caption{AMCL}
\label{fig:3d_map_amcl}
\end{subfigure}
\begin{subfigure}{0.18\textwidth}
\centering
\includegraphics[width=0.82\textwidth]{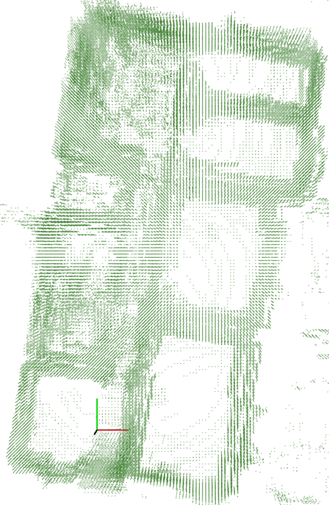}
\caption{UKFL}
\end{subfigure}
\begin{subfigure}{0.18\textwidth}
\centering
\includegraphics[width=0.7\textwidth]{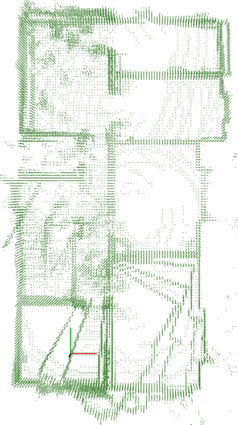}
\caption{Cartographer}
\label{fig:3d_map_amcl}
\end{subfigure}
\label{fig:3d_map_ukfl}
\begin{subfigure}{0.18\textwidth}
\centering
\includegraphics[width=0.75\textwidth]{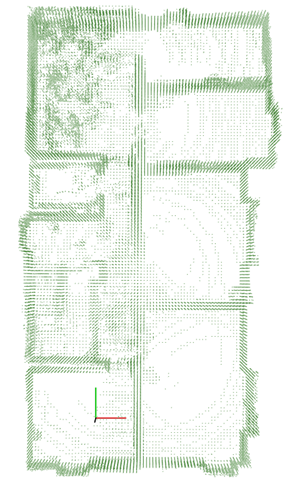}
\caption{\textit{iS-Graphs}}
\label{fig:3d_map_is_graphs}
\end{subfigure}
\caption{3D maps generated using robot poses for real-world sequence D2. (a) Ground truth 3D map from BIM. (b) 3D map generated by AMCL. (c) 3D map generated by UKFL \cite{hdl_graph_slam}. (d) 3D map generated by Cartographer. (e) 3D map generated by our \textit{iS-Graphs}.}
\label{fig:3D_maps}
\end{figure*}

\subsection{Results and Discussion}

\textbf{Simulated Datasets.}
Table.~\ref{tab:ape_simulation} presents the APE of our proposed \textit{iS-Graphs} and state-of-the LiDAR-based localization algorithms for six simulated datasets. Each simulated dataset represents a single floor level of a given construction environment with varying configurations of wall-surfaces and rooms. Given the output from 2D LiDAR algorithms, the APE of all the algorithms is computed in 2D ($x$, $y$, $\theta$). Figure \ref{fig:ape_simulation} shows the top view of estimated trajectories in D2 sequence for our algorithm and all the baselines. It can be seen from Table.~\ref{tab:ape_simulation} that our \textit{iS-Graphs} show higher robustness against localization failure and outperform all the localization baselines using both 2D and 3D LiDARs. 

%We tested and evaluated our proposed approach in various datasets created from BIM models, and compared its performance with state-of-the-art algorithms. The evaluation was based on computing the absolute pose error(APE) in the estimated location $(x \& y)$ and direction $\theta$ of the robot w.r.t the ground truth. Since AMCL is a 2D-localization algorithm, therefore in order for a fair comparison we only compare the APE in  $x \ y \& \theta$.  To ensure a fair comparison, we exclude the first 20 seconds, which represent the initialization stage for the MCL, when calculating the localization error. We consider the method to have converged when calculating the APE and RMSE, and regard localization as unsuccessful if the method fails to converge within 20 seconds, as determined by a location RMSE or yaw RMSE exceeding a predefined threshold. The results of the experiments are shown in \ref{tab:ape_simulation}. 

%It is evident that our hierarchical graph matching-based localization method outperforms traditional localization approaches significantly. It is worth noting that the environments used to validate our approach and compare it to other baselines were highly symmetric. As a result, traditional algorithms such as AMCL and UKF localization were unable to accurately localize the robot. Our localization approach only experiences difficulty when the robot is unable to detect a room as designed in the A-Graph.

\textbf{Real-World Datasets.}
%To further validate the potential of the proposed \textit{iS-Graphs}, we validate it over real datasets from construction environments.
%In real-world datasets, we compare the point cloud RMSE of the 3D maps generated using the pose estimated by the localization algorithms with the ground truth point cloud from the architectural plans.
Table.~\ref{tab:pcerror_real_data} presents the point cloud RMSE for three different construction environments, where our \textit{iS-Graphs} is able to localize the robot correctly while providing a more accurate 3D map of the environment when compared to the ground truth. Given the presence of noise in LiDAR measurements and the clutter present in a real construction environment traditional approaches to localization fail to localize the robot (see D1 in Table.~\ref{tab:pcerror_real_data}).
Since our approach relies only on high-level entities like wall-surfaces and rooms and their topological relationship instead of directly relying on low-level LiDAR measurements, it is more robust to noise and clutter in the environment than the baselines, robustly localizing the robot in all the real-world datasets and achieving the lowest point cloud RMSE. Additionally, Fig~\ref{fig:3D_maps} shows the qualitative results of the aligned map generated using robot poses obtained from \textit{iS-Graphs} and the baselines for D2 against the ground truth map generated using BIM. \textit{iS-Graphs} shows accurate alignment with the ground truth map (see axis in Fig.~\ref{fig:3d_map_gt} and Fig.~\ref{fig:3d_map_is_graphs}), while AMCL, UKFL, and Cartographer due to inaccuracies in localization provide inaccurate robot pose and hence poorly aligned and noisier map when comparing it against the ground truth.

\subsection{Limitations}
Although \textit{iS-Graphs} has demonstrated state-of-the-art performance in our experimental validation, there are still some open challenges. The success of our graph-matching algorithm depends on the correct extraction of rooms in the \textit{S-Graphs}. In cases where \textit{S-Graph} fails to detect a room that is present in the BIM, our graph matching may encounter difficulties in finding a reliable match.

Additionally, the success of \textit{iS-Graphs} is reliant on the ability to obtain a unique match between \textit{S-Graph} and \textit{A-Graph}. 
In cases where the environment has high symmetry, the graph-matching module will not be able to find a unique match candidate until, at least, one unique combination of rooms is detected by \textit{S-Graphs}.
\section{Conclusion}
\label{conclusion}
In this paper, we have presented a novel method for global robot localization utilizing prior information from architectural plans. We propose to embed the architectural data from BIM models into optimizable graphs that we denote as Architectural Graphs (\textit{A-Graphs}). Utilizing Situational Graphs (\textit{S-Graphs}) estimated by a robot as it navigates its environment, we present a novel graph matching strategy to match the \textit{A-Graphs} with the \textit{S-Graphs} and we also present a graph merging strategy to fuse the content of both. The result of the fusion is an informed \textit{iS-Graph}, which enables the robot to localize itself within the architectural plan for the given environment. We validate our approach on different simulated and real-world construction sites showcasing state-of-the-art results with respect to current localization frameworks. In future works, we will explore the possibility of incorporating uncertainties in the architectural plans to further detect and correct deviations in the built environment with respect to the architectural plans.     

%augmenting the situational graphs with prior knowledge, and localization of mobile robots by matching the prior and online situational graphs of the environment. 
\balance
\bibliographystyle{IEEEtran}
\bibliography{Biobliography}

\begin{thebibliography}{10}
\providecommand{\url}[1]{#1}
\csname url@rmstyle\endcsname
\providecommand{\newblock}{\relax}
\providecommand{\bibinfo}[2]{#2}
\providecommand\BIBentrySTDinterwordspacing{\spaceskip=0pt\relax}
\providecommand\BIBentryALTinterwordstretchfactor{4}
\providecommand\BIBentryALTinterwordspacing{\spaceskip=\fontdimen2\font plus
\BIBentryALTinterwordstretchfactor\fontdimen3\font minus
  \fontdimen4\font\relax}
\providecommand\BIBforeignlanguage[2]{{%
\expandafter\ifx\csname l@#1\endcsname\relax
\typeout{** WARNING: IEEEtran.bst: No hyphenation pattern has been}%
\typeout{** loaded for the language `#1'. Using the pattern for}%
\typeout{** the default language instead.}%
\else
\language=\csname l@#1\endcsname
\fi
#2}}

\bibitem{bim}
S.~Azhar, M.~Hein, and B.~Sketo, ``Building information modeling (bim):
  benefits, risks and challenges,'' in \emph{Proceedings of the 44th ASC Annual
  Conference}, 2008, pp. 2--5.

\bibitem{connecting_semantic_bim}
R.~W.~M. Hendrikx, P.~Pauwels, E.~Torta, H.~J. Bruyninckx, and M.~J.~G. van~de
  Molengraft, ``Connecting semantic building information models and robotics:
  An application to 2d lidar-based localization,'' in \emph{2021 IEEE
  International Conference on Robotics and Automation (ICRA)}, 2021, pp.
  11\,654--11\,660.

\bibitem{bim_localization}
M.~S. Moura, C.~Rizzo, and D.~Serrano, ``Bim-based localization and mapping for
  mobile robots in construction,'' in \emph{2021 IEEE International Conference
  on Autonomous Robot Systems and Competitions (ICARSC)}, 2021, pp. 12--18.

\bibitem{robot_localization_shaheer}
\BIBentryALTinterwordspacing
M.~Shaheer, H.~Bavle, J.~L. Sanchez-Lopez, and H.~Voos, ``Robot localization
  using situational graphs and building architectural plans,'' 2022. [Online].
  Available: \url{https://arxiv.org/abs/2209.11575}
\BIBentrySTDinterwordspacing

\bibitem{s_graphs}
H.~Bavle, J.~L. Sanchez-Lopez, M.~Shaheer, J.~Civera, and H.~Voos,
  ``Situational graphs for robot navigation in structured indoor
  environments,'' \emph{IEEE Robotics and Automation Letters}, vol.~7, no.~4,
  pp. 9107--9114, 2022.

\bibitem{s_graphs+}
\BIBentryALTinterwordspacing
------, ``S-graphs+: Real-time localization and mapping leveraging hierarchical
  representations,'' 2022. [Online]. Available:
  \url{https://arxiv.org/abs/2212.11770}
\BIBentrySTDinterwordspacing

\bibitem{majdik2013mav}
A.~L. Majdik, Y.~Albers-Schoenberg, and D.~Scaramuzza, ``Mav urban localization
  from google street view data,'' in \emph{2013 IEEE/RSJ International
  Conference on Intelligent Robots and Systems}.\hskip 1em plus 0.5em minus
  0.4em\relax IEEE, 2013, pp. 3979--3986.

\bibitem{kim2017satellite}
D.-K. Kim and M.~R. Walter, ``Satellite image-based localization via learned
  embeddings,'' in \emph{2017 IEEE International Conference on Robotics and
  Automation (ICRA)}.\hskip 1em plus 0.5em minus 0.4em\relax IEEE, 2017, pp.
  2073--2080.

\bibitem{kendall2016modelling}
A.~Kendall and R.~Cipolla, ``Modelling uncertainty in deep learning for camera
  relocalization,'' in \emph{2016 IEEE international conference on Robotics and
  Automation (ICRA)}.\hskip 1em plus 0.5em minus 0.4em\relax IEEE, 2016, pp.
  4762--4769.

\bibitem{lowry2015visual}
S.~Lowry, N.~S{\"u}nderhauf, P.~Newman, J.~J. Leonard, D.~Cox, P.~Corke, and
  M.~J. Milford, ``Visual place recognition: A survey,'' \emph{ieee
  transactions on robotics}, vol.~32, no.~1, pp. 1--19, 2015.

\bibitem{desrochers2015set}
B.~Desrochers, S.~Lacroix, and L.~Jaulin, ``Set-membership approach to the
  kidnapped robot problem,'' in \emph{2015 IEEE/RSJ International Conference on
  Intelligent Robots and Systems (IROS)}.\hskip 1em plus 0.5em minus
  0.4em\relax IEEE, 2015, pp. 3715--3720.

\bibitem{miller2021any}
I.~D. Miller, A.~Cowley, R.~Konkimalla, S.~S. Shivakumar, T.~Nguyen, T.~Smith,
  C.~J. Taylor, and V.~Kumar, ``Any way you look at it: Semantic crossview
  localization and mapping with lidar,'' \emph{IEEE Robotics and Automation
  Letters}, vol.~6, no.~2, pp. 2397--2404, 2021.

\bibitem{leung2008localization}
K.~Y.~K. Leung, C.~M. Clark, and J.~P. Huissoon, ``Localization in urban
  environments by matching ground level video images with an aerial image,'' in
  \emph{2008 IEEE International Conference on Robotics and Automation}.\hskip
  1em plus 0.5em minus 0.4em\relax IEEE, 2008, pp. 551--556.

\bibitem{zuo2020multimodal}
X.~Zuo, W.~Ye, Y.~Yang, R.~Zheng, T.~Vidal-Calleja, G.~Huang, and Y.~Liu,
  ``Multimodal localization: Stereo over lidar map,'' \emph{Journal of Field
  Robotics}, vol.~37, no.~6, pp. 1003--1026, 2020.

\bibitem{Markov-Loc}
D.~Fox, W.~Burgard, and S.~Thrun, ``{Markov Localization for Reliable Robot
  Navigation and People Detection},'' \emph{Journal of Artificial Intelligence
  Research}, vol.~11, pp. 391--427, 1999.

\bibitem{MCL}
F.~Dellaert, D.~Fox, S.~Thrun, and W.~Burgard, ``{Monte Carlo Localization for
  Mobile Robots},'' 1999.

\bibitem{scan-MCL}
J.~Röwekämper, C.~Sprunk, G.~D. Tipaldi, C.~Stachniss, P.~Pfaff, and
  W.~Burgard, ``{On the position accuracy of mobile robot localization based on
  particle filters combined with scan matching},'' 1999.

\bibitem{mpd-MCL}
S.~Thrun, D.~Fox, and W.~Burgard, ``{Monte Carlo localization with mixture
  proposal distribution},'' 2000, pp. 859--865.

\bibitem{ndt-mcl}
J.~Saarinen, H.~Andreasson, T.~Stoyanov, and A.~J. Lilienthal, ``{Normal
  Distributions Transform Monte-Carlo Localization (NDT-MCL)},''
  \emph{International Conference on Intelligent Robots and Systems (IROS)},
  2013.

\bibitem{cartographer}
W.~Hess, D.~Kohler, H.~Rapp, and D.~Andor, ``Real-time loop closure in 2d lidar
  slam,'' in \emph{2016 IEEE International Conference on Robotics and
  Automation (ICRA)}, 2016, pp. 1271--1278.

\bibitem{fpfh}
R.~Rusu, N.~Blodow, and M.~Beetz, ``{Fast point feature histograms (fpfh) for
  3d registration},'' 2009.

\bibitem{sog}
F.~Tombari, S.~Salti, and L.~D. Stefano, ``{Unique signatures of histograms for
  local surface description},'' 2010.

\bibitem{bowpf}
B.~Steder, M.~Ruhnke, S.~Grzonka, and W.~Burgard, ``{Place recognition in 3d
  scans using a combination of bag of words and point feature based relative
  pose estimation},'' 2011.

\bibitem{fast_histo}
T.~Rhling, J.~Mack, and D.~Schulz, ``{fast histogram-based similarity measure
  for detecting loop closures in 3-d lidar data},'' 2015.

\bibitem{M2dp}
L.~He, X.~Wang, and H.~Zhang, ``{M2dp: A novel 3d point cloud descriptor and
  its application in loop closure detection},'' 2016.

\bibitem{seg-map}
R.~Dubé, A.~Cramariuc, D.~Dugas, H.~Sommer, M.~Dymczyk, J.~Nieto, R.~Siegwart,
  and C.~Cadena, ``Segmap: Segment-based mapping and localization using
  data-driven descriptors,'' \emph{The International Journal of Robotics
  Research}, vol.~39, no. 2-3, pp. 339--355, 2020.

\bibitem{PGLOC-CADfloorPlans}
F.~Boniardi, T.~Caselitz, R.~Kümmerle, and W.~Burgard, ``{A pose graph-based
  localization system for long-term navigation in CAD floor plans},''
  \emph{Robotics and Autonomous Systems}, vol. 112, 2019.

\bibitem{room-layout-edge-extraction}
F.~Boniardi, A.~Valada, R.~Mohan, T.~Caselitz, and W.~Burgard, ``{Robot
  localization in floor plans using a room layout edge extraction network},''
  \emph{International Conference on Intelligent Robots and Systems (IROS)},
  2019.

\bibitem{GLFP}
X.~Wang, R.~J.~Marcotte, and E.~Olson, ``{GLFP: Global Localization from a
  Floor Plan},'' \emph{International Conference on Intelligent Robots and
  Systems (IROS)}, 2019.

\bibitem{Precise-Robot-Localization-in-Architectural-3D-Plans}
H.~Blum, J.~Stiefel, C.~Cadena, R.~Siegwart, and A.~Gawel, ``{Precise Robot
  Localization in Architectural 3D Plans},'' \emph{arXiv preprint
  arXiv:2006.05137}, 2020.

\bibitem{SGD-LOC}
Z.~Li, M.~H., A.~Jr, and D.~Rus, ``{Online Localization with Imprecise Floor
  Space Maps using Stochastic Gradient Descent},'' \emph{International
  Conference on Intelligent Robots and Systems (IROS)}, 2020.

\bibitem{AMCL}
D.~Fox, ``Kld-sampling: Adaptive particle filters,'' \emph{Advances in neural
  information processing systems}, vol.~14, 2001.

\bibitem{FP-LOC}
L.~Gao and K.~Laurent, ``{FP-Loc: Lightweight and Drift-free Floor
  Plan-assisted LiDAR Localization},'' \emph{International Conference on
  Robotics and Automation (ICRA)}, 2022.

\bibitem{Towards-BIM-based-robot-localization}
H.~Yin, J.~Liew, W.~Lee, M.~Ang~Jr, and J.~Yeoh, ``{Towards BIM-based robot
  localization: a real-world case study},'' \emph{International Symposium on
  Automation and Robotics in Construction}, 2022.

\bibitem{vega:2022:2DLidarLocalization}
\BIBentryALTinterwordspacing
M.~Vega~Torres, A.~Braun, and A.~Borrmann, ``Occupancy grid map to pose
  graph-based map: Robust bim-based 2d- lidar localization for lifelong indoor
  navigation in changing and dynamic environments,'' in \emph{eWork and
  eBusiness in Architecture, Engineering and Construction: ECPPM 2022},
  S.~F.~S. Eilif~Hjelseth and R.~Scherer, Eds.\hskip 1em plus 0.5em minus
  0.4em\relax CRC Press, Sep 2022. [Online]. Available:
  \url{https://publications.cms.bgu.tum.de/2022_ECPPM_Vega.pdf}
\BIBentrySTDinterwordspacing

\bibitem{BIM-LOC}
Y.~Huan, L.~Zhiyi, and Y.~K.W.Justin, ``{Semantic localization on BIM-generated
  maps using a 3D LiDAR sensor},'' \emph{Automation in Construction}, vol. 146,
  2023.

\bibitem{bailey2000data}
T.~Bailey, E.~M. Nebot, J.~Rosenblatt, and H.~F. Durrant-Whyte, ``Data
  association for mobile robot navigation: A graph theoretic approach,'' in
  \emph{Proceedings 2000 ICRA. Millennium Conference. IEEE International
  Conference on Robotics and Automation. Symposia Proceedings (Cat. No.
  00CH37065)}, vol.~3.\hskip 1em plus 0.5em minus 0.4em\relax IEEE, 2000, pp.
  2512--2517.

\bibitem{antonante2021outlier}
P.~Antonante, V.~Tzoumas, H.~Yang, and L.~Carlone, ``Outlier-robust estimation:
  Hardness, minimally tuned algorithms, and applications,'' \emph{IEEE
  Transactions on Robotics}, vol.~38, no.~1, pp. 281--301, 2021.

\bibitem{lusk2021clipper}
P.~C. Lusk, K.~Fathian, and J.~P. How, ``Clipper: A graph-theoretic framework
  for robust data association,'' in \emph{2021 IEEE International Conference on
  Robotics and Automation (ICRA)}.\hskip 1em plus 0.5em minus 0.4em\relax IEEE,
  2021, pp. 13\,828--13\,834.

\bibitem{arandjelovic2016netvlad}
R.~Arandjelovic, P.~Gronat, A.~Torii, T.~Pajdla, and J.~Sivic, ``Netvlad: Cnn
  architecture for weakly supervised place recognition,'' in \emph{Proceedings
  of the IEEE conference on computer vision and pattern recognition}, 2016, pp.
  5297--5307.

\bibitem{global_loc_gnn}
Y.~Liu, Y.~Petillot, D.~Lane, and S.~Wang, ``Global localization with
  object-level semantics and topology,'' in \emph{2019 International Conference
  on Robotics and Automation (ICRA)}, 2019, pp. 4909--4915.

\bibitem{semantic_loop_closure}
\BIBentryALTinterwordspacing
C.~Qin, Y.~Zhang, Y.~Liu, and G.~Lv, ``Semantic loop closure detection based on
  graph matching in multi-objects scenes,'' \emph{Journal of Visual
  Communication and Image Representation}, vol.~76, p. 103072, 2021. [Online].
  Available:
  \url{https://www.sciencedirect.com/science/article/pii/S1047320321000389}
\BIBentrySTDinterwordspacing

\bibitem{zheng2020buildingfusion}
T.~Zheng, G.~Zhang, L.~Han, L.~Xu, and L.~Fang, ``Buildingfusion:
  Semantic-aware structural building-scale 3d reconstruction,'' \emph{IEEE
  Transactions on Pattern Analysis and Machine Intelligence}, vol.~44, no.~5,
  pp. 2328--2345, 2020.

\bibitem{scene_graph}
U.~Kim, J.~Park, T.~Song, and J.~Kim, ``{3-D Scene Graph: A Sparse and Semantic
  Representation of Physical Environments for Intelligent Agents},'' in
  \emph{IEEE TRANSACTIONS ON CYBERNETICS}, 2019, pp. 1–--13.

\bibitem{3d_scene_graph}
I.~Armeni, Z.-Y. He, J.~Gwak, A.~R. Zamir, M.~Fischer, J.~Malik, and
  S.~Savarese, ``{3D Scene Graph: A structure for unified semantics, 3D space,
  and camera},'' in \emph{Proceedings of the IEEE/CVF International Conference
  on Computer Vision}, 2019, pp. 5664--5673.

\bibitem{hydra}
N.~Hughes, Y.~Chang, and L.~Carlone, ``{Hydra: A Real-time Spatial Perception
  Engine for 3D Scene Graph Construction and Optimization},'' \emph{arXiv
  preprint arXiv:2201.13360}, 2022.

\bibitem{hdl_graph_slam}
K.~Koide, J.~Miura, and E.~Menegatti, ``A portable three-dimensional
  {LIDAR}-based system for long-term and wide-area people behavior
  measurement,'' \emph{International Journal of Advanced Robotic Systems},
  vol.~16, no.~2, Mar. 2019.

\bibitem{evo_traj_calc}
M.~Grupp, ``{evo: Python package for the evaluation of odometry and SLAM},''
  \url{https://github.com/MichaelGrupp/evo}, 2017.

\end{thebibliography}

\end{document}